# Vision-based module for accurately reading linear scales in a laboratory

Parvesh Saini, Soumyadipta Maiti, Beena Rai

*Abstract*— Capabilities and the number of vision-based models are increasing rapidly. And these vision models are now able to do more tasks like object detection, image classification, instance segmentation etc. with great accuracy. But models which can take accurate quantitative measurements form an image, as a human can do by just looking at it, are rare. For a robot to work with complete autonomy in a Laboratory environment, it needs to have some basic skills like navigation, handling objects, preparing samples etc. to match human-like capabilities in an unstructured environment. Another important capability is to read measurements from instruments and apparatus.

Here, we tried to mimic a human inspired approach to read measurements from a linear scale. As a test case we have picked reading level from a syringe and a measuring cylinder. For a randomly oriented syringe we carry out transformations to correct the orientation. To make the system efficient and robust, the area of interest is reduced to just the linear scale containing part of the image. After that, a series of features were extracted like the major makers, the corresponding digits, and the level indicator location, from which the final reading was calculated. Readings obtained using this system were also compared against human read values of the same instances and an accurate correspondence was observed.

## I. INTRODUCTION

The effects of advancement in artificial intelligence and machine learning can be observed in many industries such as Manufacturing, Healthcare, Finance etc. [1]. This has led to major boosts particularly in areas like autonomous vehicles [2], industrial robots [3], healthcare [4], agriculture [5], autonomous drones [6], and smart cities [7]. Because of the ability of these AI/ML models to find trends in large multidimensional datasets, they are actively being adopted for the tasks like clinical diagnostics [8], drug discovery [9, 10] and chemical formulation development [11] which conventionally required human expertise. Growth in these sectors was hampered due to the human limitations of not being able to perceive high-dimensional features at scale and speed.

For these ML recommender models to work satisfactorily, a huge amount of data or experimentation is needed. And if one wishes to use those models for tasks involving frequent iterations and experimentation like chemical formulation discovery, lots of experimental trials need to be done to build a big dataset for formulations and their respective physical and chemical properties. But it is practically very difficult for a human to do these many trial experiments in a fast manner. Thus, to accelerate this, humans need to be assisted or replaced by a robot in a laboratory [12].

The idea of having a robot working in a lab and carrying out experiments is not new. Many attempts have been made such as [13] a mobile robot chemist and demonstrated robot working in a lab for 8 days and carrying out 688 experiments. That lab environment was structured specifically for the robot with calibration cubes placed in front of each equipment for recalibration. However, a robot working with full autonomy like a human being in an unstructured lab environment is still rare. For this purpose, there are many individual capabilities required like navigation in the lab, operating existing equipment, handling glass apparatus etc. to achieve full autonomy.

Here in this paper, we have focused on the perception capability of the robot. As the laboratory environment is filled with apparatus, equipment, and devices with scales on them, a robot working on it needs to be able to read these scales. Some existing works are there in the literature for reading circular scales like dials [14, 15] but attempts to accurately read linear scales are limited [16, 17]. We aimed to have a linear scale in an unstructured environment and the robot should be able to read it from an arbitrary distance from the scale with any random angle in the frame. Like a human being, this capability should be independent of the type of laboratory apparatus like syringe, pipette or burette being used and the vision module should be able to give reading directly by looking at the markings on the scale.

To achieve this, we have developed a pipeline with multiple steps of image processing and feature extraction from the frame which is robust enough to work with various orientations of the scale and distances from the camera.

This paragraph of the first footnote will contain the date on which you submitted your paper for review, which is populated by IEEE. It is IEEE style to display support information, including sponsor and financial support acknowledgment, here and not in an acknowledgment section at the end of the article. *This work received no funding.* The name of the corresponding author appears after the financial information, e.g., **(Corresponding author: Parvesh Saini)**. Parvesh Saini is the first author.

Parvesh Saini, Soumyadipta Maitia, and Beena Raia are with TCS Research, Tata Consultancy Services Limited, Hinjewadi Phase III, Pune 411057, Maharashtra, India (e-mail: parvesh.10@tcs.com).

## II. METHODOLOGY

This developed AI pipeline starts with the image capture, either as a frame of a video/live feed or as a pre-captured image. The steps involved are Object Detection, Orientation Correction, Feature Extraction and Level Calculation using the level indicator for the apparatus. A python script was developed for this purpose using YOLO [18] and OpenCV [19] modules like Find contours, Tesseract, and color segmentation which can process the frame, extract features, and give a reading for the scale in the frame. These individual steps are discussed in detail below, shown in Fig. 1.

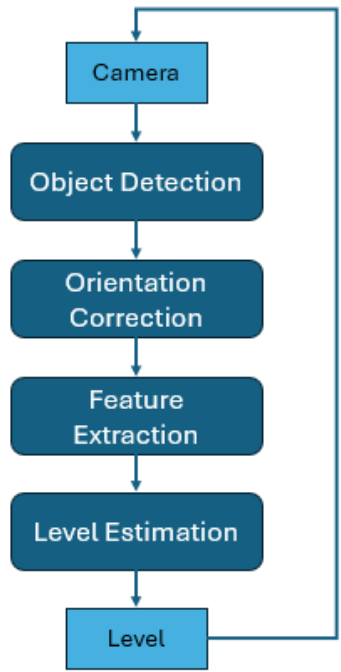

*Figure 1: Flow of the steps involved*

### A. Image Processing & Object Detection

The image from the source can be read directly in the python environment using OpenCV's "imread()" functionality and it creates a 3-dimensional array corresponding to BGR color channels of the image.

It might not always be possible to just capture the scale in the frame and some extra objects might get captured. These objects, along with taking the processing effort, can also interfere with the various feature extraction steps down the pipeline. Therefore, it would make sense to crop into the part of the image relevant for us i.e., the object with a scale. To achieve this, we have used the YOLO package for object detection which is a model pretrained on large object classes and can be modified to detect custom objects.

Now, this fine-tuned YOLO model for syringe is loaded and image is passed through the model. YOLO will give various instances for each object in the finetuning (Linear Scale). For each instance, co-ordinate of an enclosing rectangle can be extracted and a cropped image for that instance can be obtained.

But YOLO only gives you a box enclosing the object and not the orientation. It is critical for the object to be oriented correctly because the OCR module works best when the marker characters are vertical.

### B. Orientation Correction

To correct the orientation of the image, we first need to define the features with respect to which orientation of the object will be defined. Here, for linear scales these features can be assumed as the markings on the scale, the digits and the level markers.

These scale orientation related features are extracted using color-based segmentation as these markers are usually in contrast with the surrounding area for better visibility. Segmentation results in a binary mask of the image in which the selected parts are in white and the rest of the image is black. This binary image is then used to extract contour around these selected parts, using the "findContour()" method in OpenCV. These contours are nothing but co-ordinates on the periphery of the binary features along with some of their properties like area and perimeter. A separate exercise of contour analysis can be carried out to determine tight area and perimeter bounds to filter these orienting features. Using tight values of these properties we can further narrow down the contours detected to just the contours defining the orientation of the object.

Now that we are just left with the contours (collection of points), we can use Principal Component Analysis (PCA) to extract major and minor axes along with their angles for the collection of points corresponding to the orienting contours. We can thus rotate the image and the points in the opposite direction with the same angles and get the oriented image. To further narrow down the area of interest in the images we can use these now correctly oriented features to crop up to just the scale part of the object and carry further analysis. The flow of the steps mentioned above is represented in Fig. 2.

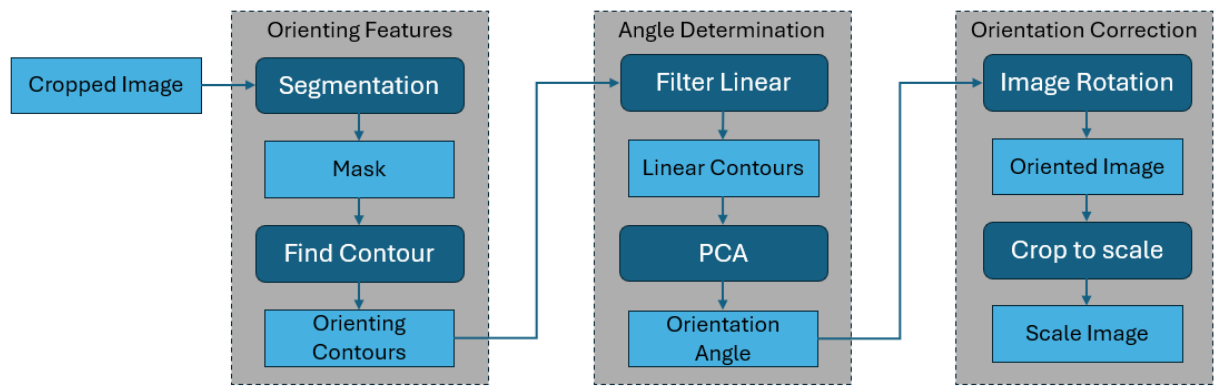

*Figure 2: Flow of steps to correct the orientation of the object and cropping to scale*

*C. Feature Extraction*

As we have now eliminated the irrelevant part of the image and cropped to just the oriented scale in the frame, we can proceed further with feature extraction steps that would be necessary to get a reading from this image. To make any sense out of an array of pixels (an image) some features need to be extracted which are relevant for the current application. The features relevant to us are the level markers, level digits and the level indicator. But before that a step of pre-processing needs to be carried out on the image to make it suitable for the feature extraction steps. The series of steps from pre-processing to level estimation can be seen in Fig. 3.

artificial blurring to the image. This is achieved using Gaussian Blur functionality in OpenCV using a 3x3 kernel where central value is the average of all other values in the kernel.

*Marker Extraction*

As the level makers on a scale are always present in a distinct shade from the background for a clear visibility for the user, color-based segmentation can be used to separate out these elements from the image. After segmentation of the image, most of the information is discarded as it gets converted into a binary image. The parts with pixel values that lie within the color range are white (value 255) and the rest is black (value 0).

The binary image obtained is fed to a contour detection module "findContours()" present in OpenCV. The module extracts all the closed contours present in that binary image along with its properties like area and parameter which can be used for further analysis. Once all the contours are obtained, we can filter contours based on their size and aspect ratio

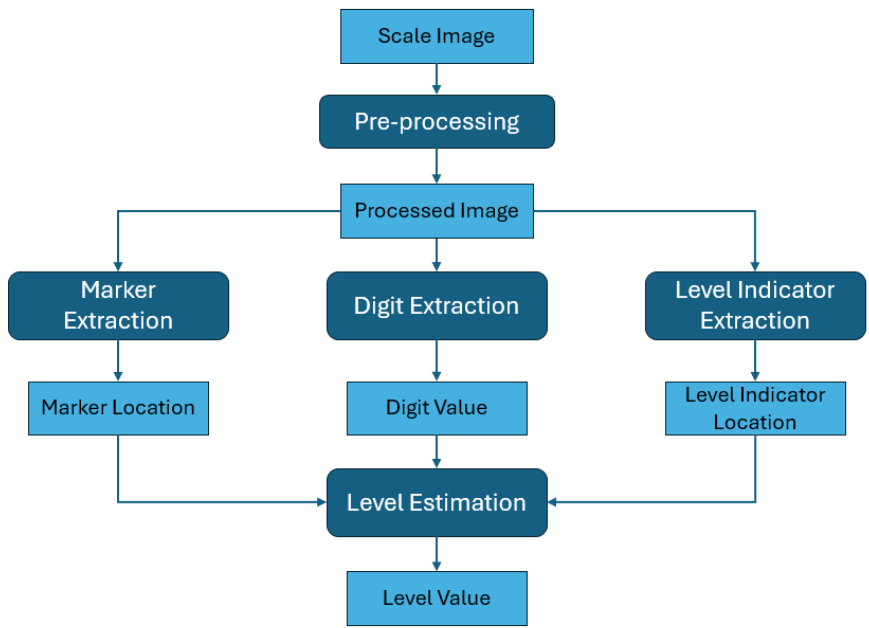

Figure 3: Feature extractions steps involved in level estimation

*Pre-Processing*

The first pre-processing which is applied to the image is change of color space. By default, OpenCV reads an image in BGR color space which means that for a pixel the intensity values are written in blue, green and red. But separating out the parts of the image based on the pixel color was found to be more suitable in the HSV color space i.e., hue, saturation, and value. Thus, the image is converted to HSV color space using "cvtColor()" function of OpenCV.

Next, to eliminate noise and small features in the image which can disrupt the function of further feature extraction techniques (like edge detection), are eliminated by applying values. All the contours are subjected to a size check where the contours lying outside a perimeter and an area range (determined separately) are removed. Then, for the remaining contours their aspect ratio is determined using the extreme values of each co-ordinate.

$$aspect\ ratio = \frac{x_{max} - x_{min}}{y_{max} - y_{min}} \quad (1)$$

Only the contours above an aspect ratio threshold of 2.5 are regarded as linear contours and stored for later use.

*Digit Extraction*

The markers themselves cannot provide any useful information unless a level value is associated with them. These values are normally written along with the major markers on a scale. Thus, we can scan a small area next to each marker to read the corresponding digits. A crop of this small area next to the marker is passed through an Optical Character Reader (OCR) module called PyTesseract. It then processes the image and gives a string value for all the characters it can find in the image. This value can be processed and converted to an integer to get stored along with the maker position.

*Level Indicator Extraction*

Next thing which is required to get the level measurement is the location of the level indicator. To get this location, we first segment the part of the cropped image which meets the color criteria as: like level markers, the level indicator also has a contrasting color to stand out. Level indicators for a syringe and a measuring cylinder can be seen in Fig. 4.

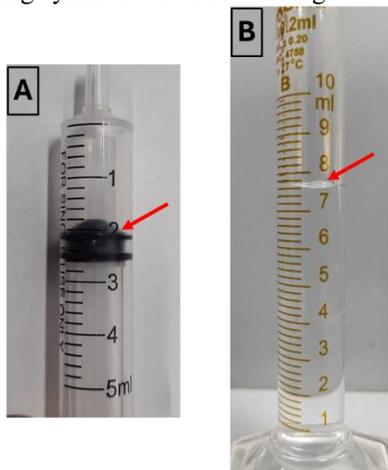

*Figure 4: Images of A) Syringe and B) Measuring Cylinder used to extract levels*

Once we have segmented the level indicator and converted it into a binary image, contour analysis can be carried out to extract contours from the segmented image. There can be a scenario where some extra contours are detected along with the level indicator, as the segmentation is not always perfect. These extra contours can then be eliminated using size bound constraints, the values of which can be determined using a separate exercise of contour analysis for level indicator.

Out of these obtained contours, which point would be regarded as the point of measurement for the level depends on the indicator. If simply the meniscus of the liquid is the level indicator as in the case of a measuring cylinder, the lowest point is set as the point of measurement or in the case of syringe the end part of the plunger touching the wall is the point of measurement.

*D. Level Estimation*

A human-inspired approach is used here as the first thing a human would also look at in an image with a scale is the level markers, and which marker corresponds to what value. Based on this, a position-to-value relation would be generated for that scale and at the end, using the position of the level indicator, the level value would be calculated.

Now that we have the position of the major markers and the level value associated with each of them, we can calculate the position-to-value relation using linear relation for the frame which can be used later to get the level from the level indicator.

$$y = mx + c \quad (2)$$

In the linear equation above, $y$ is the dependent variable, the level value in our case, and $x$ is the independent variable, the location of the measuring point on the level indicator. The other two values are slope ($m$) and $c$ (offset), which can be calculated using marker position and maker values.

III. RESULTS & DISCUSSION

Single frames were captured using a camera to test and validate the real-world performance of the system. For example, an image was captured with a syringe randomly oriented in the frame, as seen in Fig. 5, to test the robustness of the model to read values in an unstructured environment.

*A. Image Capture and Scaling*

For the first step, the raw image of randomly oriented syringe was read inside python environment. As the image can be of any arbitrary resolution it needs to be scaled to a comfortable size. For this we utilized the inbuilt function of "resize()" to scale it to a fixed size as seen in Fig. 5.

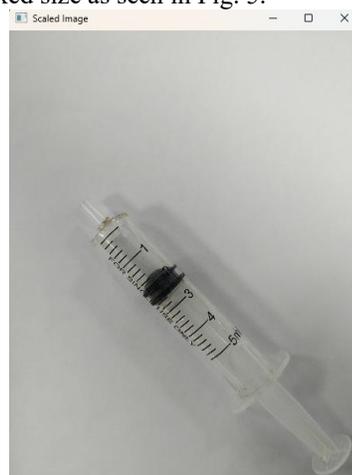

*Figure 5: Input image of a randomly oriented syringe*

*B. Object Detection*

After resizing, the image was passed through the YOLO object detection module which was fine-tuned for detecting linear scale instances. For this we collected a dataset of 80 images of linear scale instances and labelled it. After which the train functionality of YOLO was utilized to adapt to this dataset.

YOLO gives a bounding box around each object instance, as seen in Fig. 6(a). Then the image was cropped to this box, Fig. 6(b), eliminating all unnecessary parts of the image making further steps more robust.

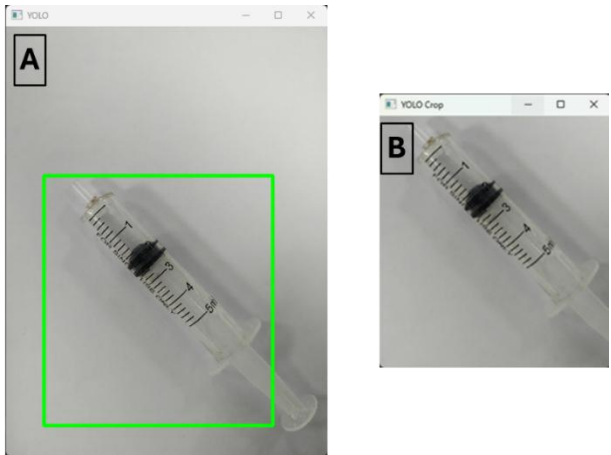

*Figure 6: A) YOLO object detection and B) cropping*

### C. Orientation Correction & Crop to Scale

The cropped version of the image needs to be reoriented so that the digits of the marker scale are vertical. To achieve this, we first need to identify and extract the features which define the orientation of the scale like the level markers. As seen in Fig. 7(a), the contours of these markers were extracted by first segmentation and masking followed by contours extraction. These contours are just a collection of points which can be used for Principal Component Analysis (PCA). PCA gave an angle of both major and minor axes corresponding to the points as seen in Fig. 7(a). Now, the image was rotated in the reverse direction of the major axis so that the syringe gets nearly vertical, as seen in Fig. 7(b).

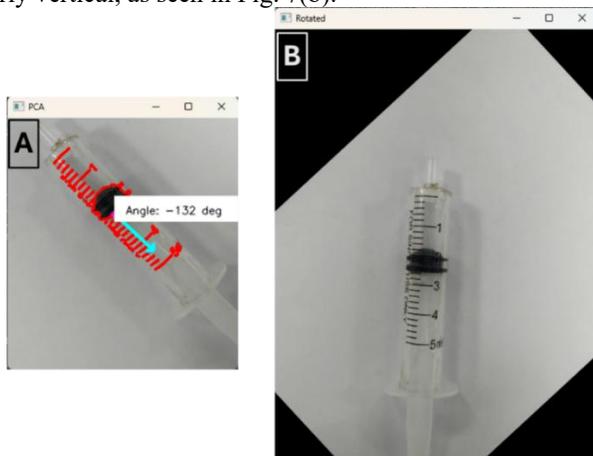

*Figure 7: A) Angle calculation using PCA for orientating features and B) Reoriented image*

To achieve even better results, one more step of orientation correction was executed. But before that, all the non-linear contours were filtered out using an aspect ratio condition where any contour with the aspect ratio (expression discussed earlier) below 2.5 is rejected and result can be seen in Fig. 8(a). To further reduce the scope of analysis in the image and make the system more robust and efficient, it was cropped down to just the part with the scale of the syringe/measuring cylinder (Fig. 8(b)). Dimensions and location of this crop were determined using max and min values of the x and y co-ordinates in the linear contours. After that a padding proportional to height and width was also applied to get the final box co-ordinates for cropping.

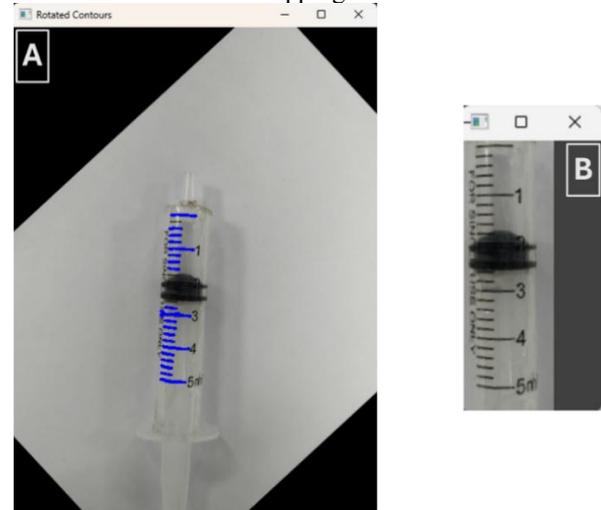

*Figure 8: A) Filtering linear contours and B) cropping to scale*

### D. Feature Extraction

As we are now left with a cropped image which only contains the scale, further feature extraction steps can be carried out. In the first step of pre-processing, Gaussian blurring using a 3x3 kernel was carried out to reduce the noise which can interfere with the feature extraction steps down the line.

*Major Contours:*

To extract major contours, we can start with the linear markers which can be obtained by segmentation, masking, contour finding and linear contour filtering, as we did before. After this, these linear contours need to be sorted into the 2 groups of major and minor markers using their relative width, as can be seen in Fig. 9. Major markers are of more importance here as the level digit is written only alongside them, which will be used later.

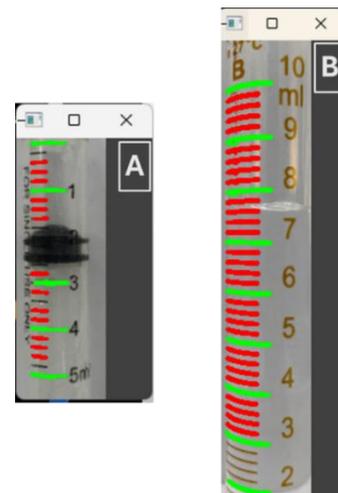

*Figure 9: Minor and major contours extracted for A) Syringe and B) Measuring Cylinder images*

Relative Contour Identifier

Depending on the distance from the camera, contours can appear to be of varied sizes in the image. So, to separate out contours of major markers, a relative classification approach was chosen. In this all the linear contours found were arranged in order of decreasing length and the jump in length of consecutive contours were calculated. Now, we need to group the contours based on the relative lengths. To do that, If the calculated jump is more than 15% of the length of the last contour, then a new group is created. This results in separate groups consisting of major and minor markers, major being the first one due to sorting. Then all the contours in the first group are passed to the next module as contours of the major markers. Successful detection of major and minor markers for two test images of different size can be seen in Fig. 10.

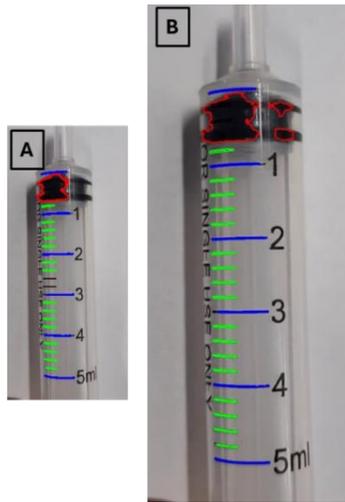

Figure 10: Major markers detected successfully for different sizes, A) 0.6 and B) 1.5 of the images (relative marker identification)

*Digits:*

After extracting major contours, level digits can be extracted. For this task we utilized the PyTesseract [20] module for Optical Character Recognition. Only the small area right next to each major marker is passed through the OCR process to improve accuracy and robustness of the system, and it also automatically associates the maker and level values, seen in Fig. 11.

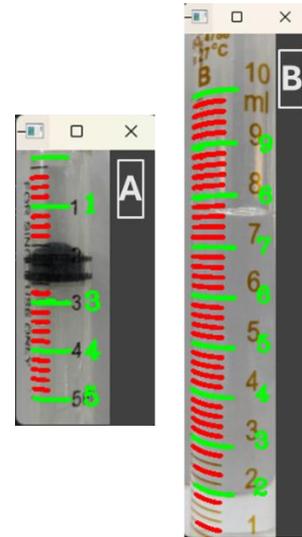

Figure 11: Reading level markers using PyTesseract OCR Module for A) Syringe and B) Measuring Cylinder

Auto-correction

Digits read by OCR are not read correctly all the time and can have misread values and lead to incorrect reading. Therefore, an auto-correction module is needed to detect and correct misread values. An example case is discussed below with initial incorrect readings shown in Table 1:

Table 1: Example with incorrect values read by OCR

| S. No | Vertical Position of detected major markers | Value (OCR) |
|---|---|---|
| 1 | 6 | Not read |
| 2 | 73 | 4 (misread) |
| 3 | 136 | 2 |
| 4 | 201 | 3 |
| 5 | 266 | 4 |

To do this, the positions and values of the digits were read and converted into pairs of all combinations and corresponding slopes were calculated for each combination (rounded off till 4 decimal point) as shown in Table 2.

Table 2: Slope calculation and analysis for different combinations

| Combination (S. No Values) | Calculated Slope | Analysis (Mode = 0.0154) |
|---|---|---|
| 2,3 | -0.0317 | Incorrect |
| 2,4 | -0.0078 | Incorrect |
| 2,5 | 0 | Incorrect |
| 3,4 | 0.0154 | Correct |
| 3,5 | 0.0154 | Correct |
| 4,5 | 0.0154 | Correct |

Point combinations which have the slope value equal to the most occurring value in the data, the mode, are regarded as correct. Using this slope and a point, the offset value of the liner expression is calculated.

$$value = slope * vertical_{position} + offset \quad (3)$$

$$offset = value - slope * vertical_{position} \quad (4)$$

$$offset = 4 - 0.0154 * 266 = -0.0964 \quad (5)$$

$$value = 0.0154 * vertical_{position} - 0.0964 \quad (6)$$

Using the linear equation obtained above, values for all the detected major contours were determined, even for the ones where OCR was not able to read anything (like Sr. No 1 value at vertical position 6). Completed set of marker position and corrected readings can be found in Table 3.

*Table 3: Corrected values for the makers*

| S. No. | Vertical Position of detected major markers | Corrected Reading |
|---|---|---|
| 1 | 6 | 0 |
| 2 | 73 | 1 |
| 3 | 136 | 2 |
| 4 | 201 | 3 |
| 5 | 266 | 4 |

The working of this Auto-correction and completion module can be seen in Fig. 12.

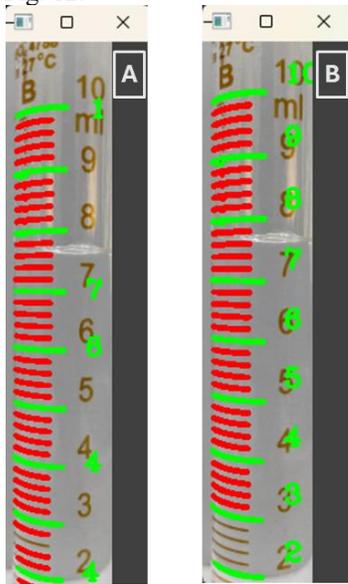

*Figure 12: Marker values A) from OCR and B) after auto-correction and completion of level values*

### Level Indicator Extraction

We now have the correct position-to-value relation. The thing that remains is the position of the level indicator in the frame. Level indicators can be of any type, here we have tested with syringe plunger plug and the meniscus in a measuring cylinder, which are one of the most common level indicators. Given their contrasting appearance, they can be extracted using segmentation, masking and contour finding techniques that we utilized earlier just with different size (For example: area between 93 and 5000) and color constraints (For example: HSV Color range between (0, 0, 0) and (255, 255, 40). Successful extraction of level indicator profile for a measuring cylinder and a syringe can be seen in Fig. 13.

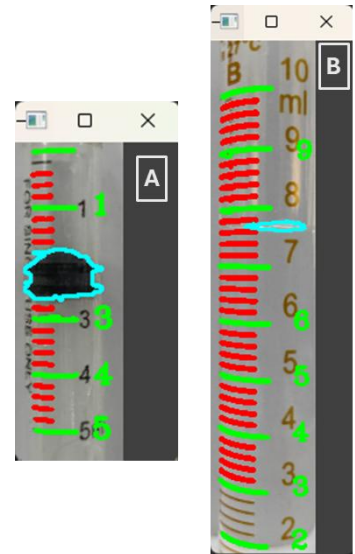

*Figure 13: Level indicator extraction for A) Syringe and B) Measuring Cylinder*

### E. Level Estimation

The point of measurement is determined from the type of liquid level indicator present. For meniscus (measuring cylinder) it will be the lowest point on the contour, and for syringe it will be at an offset from the end. The vertical position of the measurement point is then used to provide the final level value using the position-to-value linear relationship determined earlier. Different instances of level reading in the syringe and the measuring cylinder can be seen in Fig. 14. After this the pipeline was finally tested on a frame of a recorded video of a syringe with moving plunger, as can be seen in Fig. 15.

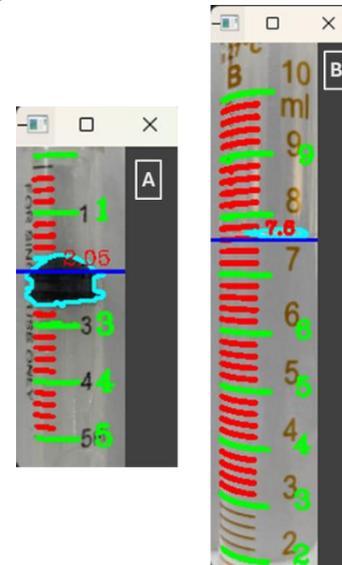

*Figure 14: Level determination for A) syringe frame and B) measuring cylinder frame*

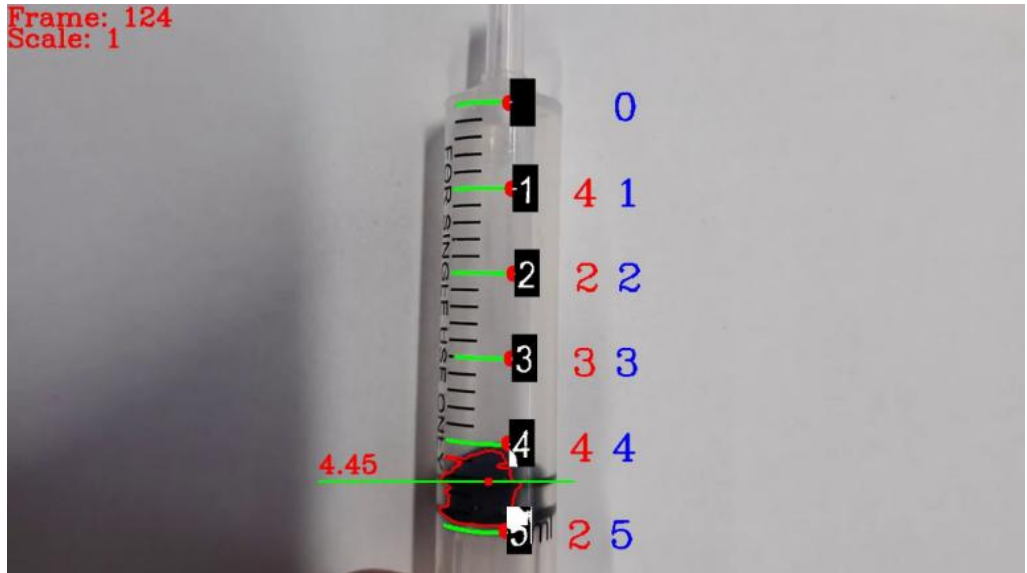
*Figure 15: Complete analysis of a syringe frame after autocorrection*

## IV. EVALUATION AND CONCLUSION

To evaluate the developed solution, measurements were made for the ground truth values where the plunger coincides with either the minor or the major markers on the scales. And to understand the effect of the direction of movement of plunger on the measurement, the values were recorded for both aspirating and dispensing action of the syringe. Values of these measurements can be found in Table 4.

*Table 4: Measurements by the vision-module for aspirating and dispensing action*

| Ground Truth / Manual measurement (ml) | CV output in Aspirating (ml) | CV output in Dispensing (ml) |
|---|---|---|
| 0.00 | 0 | 0 |
| 0.20 | 0.2 | 0.26 |
| 0.40 | 0.36 | 0.49 |
| 0.60 | 0.59 | 0.56 |
| 0.80 | 0.86 | 0.75 |
| 1.00 | 1.08 | 0.99 |
| 1.20 | 1.21 | 1.12 |
| 1.40 | 1.46 | 1.49 |
| 1.60 | 1.63 | 1.58 |
| 1.80 | 1.82 | 1.7 |
| 2.00 | 1.94 | 1.89 |
| 2.20 | 2.07 | 2.07 |
| 2.40 | 2.38 | 2.35 |
| 2.60 | 2.55 | 2.54 |
| 2.80 | 2.74 | 2.72 |
| 3.00 | 2.9 | 2.92 |
| 3.20 | 3.11 | 3.05 |
| 3.40 | 3.33 | 3.34 |
| 3.60 | 3.46 | 3.49 |
| 3.80 | 3.86 | 3.64 |
| 4.00 | 3.92 | 3.81 |
| 4.20 | 3.97 | 3.99 |
| 4.40 | 4.3 | 4.22 |
| 4.60 | 4.43 | 4.46 |

For a preliminary evaluation some common metrics like MAE, RMSE, Bias, Standard Deviation (Std. Dev.) and $R^2$ score were calculated, for the two types of movements independently, values of which can be found in Table 5.

*Table 5: Various metrics calculated for the data*

| Metric | Aspirating | Dispensing |
|---|---|---|
| MAE (ml) | 0.070 | 0.094 |
| RMSE (ml) | 0.089 | 0.109 |
| Bias (ml) | -0.043 | -0.074 |
| Std. Dev. | 0.055 | 0.055 |
| $R^2$ Score | 0.996 | 0.994 |

It can be seen that both MAE and RSME error values are low, meaning readings are accurate with low bias values as well. Small standard deviation of error values represents error in reading as consistently low irrespective of the magnitude of the measurement. High $R^2$ score (closer to 1) further indicates that the measured values have a strong linear relation with the ground truth which is required for a good measurement tool.

In addition to the metric values, some trends in the measurement were also visualized using different plots (Fig. 16) to get more insights into the data. These plots include CV 16 (d)). Values of the measurements from the two directions at the same ground truth were also plotted in Fig. 16 (e) to get the extent of agreement between the two. In the final figure

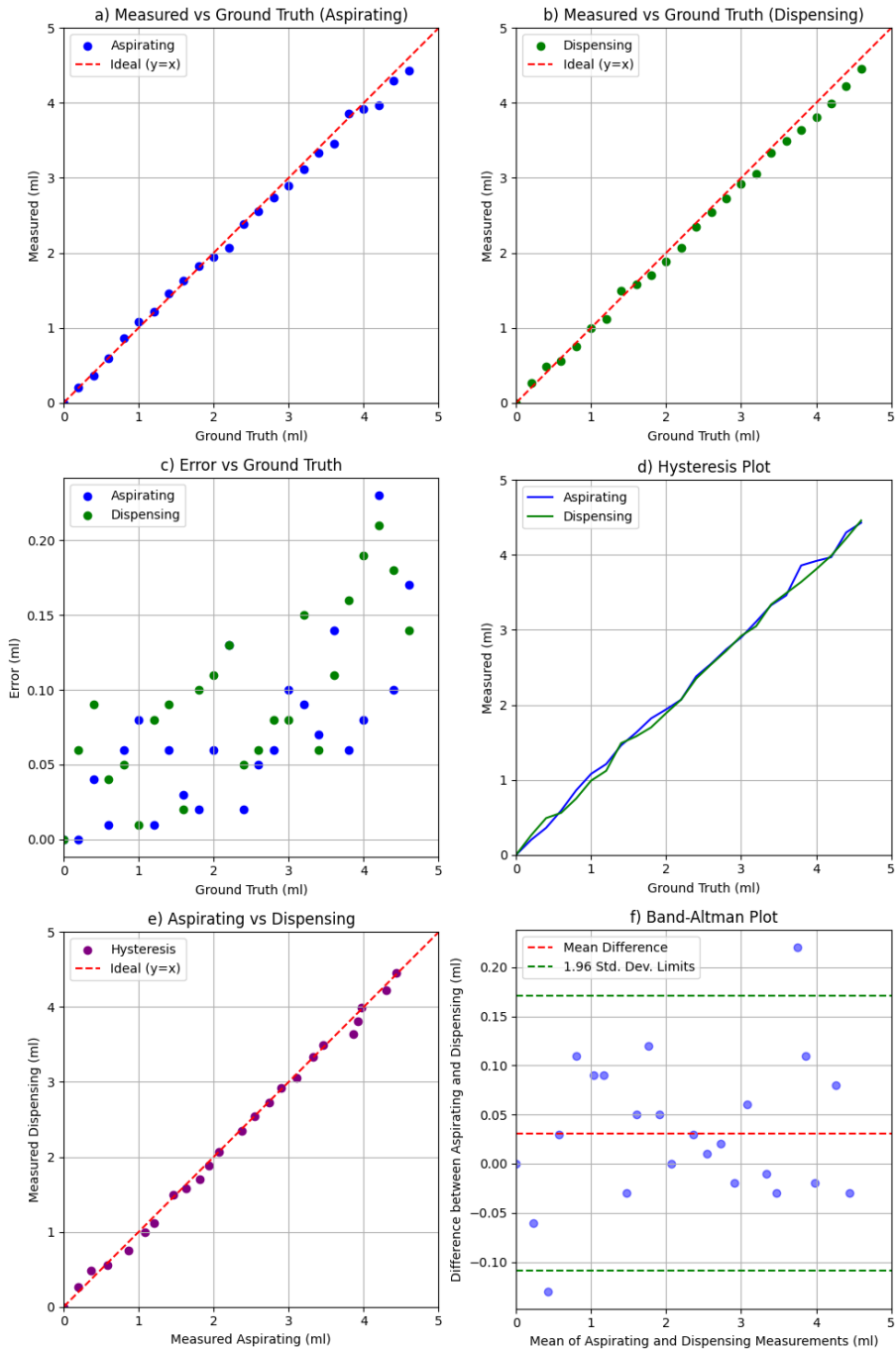

Figure 16: a) Measurements (Aspirating) vs Ground Truth, b) Measurements (Dispensing) vs Ground Truth, c) Error vs Ground Truth d) Hysteresis Curve e) Measurement (Aspirating) vs Measurement (Dispensing) f) Band-Altman plot for Aspirating and Dispensing

Measurements vs Ground Truth (Fig. 16 (a), (b)) showing how much the measured values are deviating from the ground truth. To further highlight this error of the measured reading for both aspirating and dispensing actions relative to the ground truth an Error vs Ground Truth plot is used (Fig.16 (c)). And to specifically study the effect of the direction of the plunger movement on the measurement a Hysteresis plot is used (Fig.

Band-Altman plot for aspirating and dispensing measurement is made showing spread of difference between aspirating and dispensing value around the mean.

The trend between measured and ground truth plots shows close to ideal trend between the two. The error plot hints an increasing trend with an increasing ground truth value but remains low with a maximum reaching just 0.22ml.

While analyzing the effect of direction of plunger on measurement using the hysteresis plot, the area of the loop is negligible. Also, in the Measurement (Aspirating) vs Measurement (Dispensing) plot the data points are in close agreement with each other and closely following the ideal trend. Finally, in the Band-Altman plot the spread of differences between Aspirating and Dispensing measurements are within the ±1.96 std. dev. limits of the mean difference.

Thus, based on the evidence above and by looking at the metrics and plot trends, it can be concluded that the developed Vision-based scale reading module is accurate, robust and reliable, which can be employed in any of the robotic or automation applications.

ACKNOWLEDGMENT

This research was supported by the TCS-CTO organization. We thank the TCS-RnI Infra team for providing necessary support for the execution of this work.

REFERENCES

[1] S. Makridakis, "The Forthcoming Artificial Intelligence (AI) revolution: Its Impact on Society and Firms," Futures, vol. 90, no. 90, pp. 46–60, Jun. 2017, doi: https://doi.org/10.1016/j.futures.2017.03.006
[2] C. Badue et al., "Self-driving cars: A survey," Expert Systems with Applications, vol. 165, p. 113816, Mar. 2021, doi: https://doi.org/10.1016/j.eswa.2020.113816. Available: https://www.sciencedirect.com/science/article/abs/pii/S095741742030628X
[3] R. Goel, "Role of Robotics in Health Care of the Future," Journal of Medical Academics, vol. 3, no. 1, pp. 27–29, 2020, doi: https://doi.org/10.5005/jp-journals-10070-0051
[4] F. Jiang et al., "Artificial Intelligence in healthcare: past, present and future," Stroke and Vascular Neurology, vol. 2, no. 4, pp. 230–243, Jun. 2017, doi: https://doi.org/10.1136/svn-2017-000101
[5] K. Jha, A. Doshi, P. Patel, and M. Shah, "A comprehensive review on automation in agriculture using artificial intelligence," Artificial Intelligence in Agriculture, vol. 2, no. 2, pp. 1–12, Jun. 2019, doi: https://doi.org/10.1016/j.aiia.2019.05.004
[6] D. Floreano and R. J. Wood, "Science, technology and the future of small autonomous drones," Nature, vol. 521, no. 7553, pp. 460–466, May 2015, doi: https://doi.org/10.1038/nature14542. Available: https://www.nature.com/articles/nature14542
[7] Z. Allam and Z. A. Dhunny, "On big data, artificial intelligence and smart cities," Cities, vol. 89, no. 89, pp. 80–91, Jun. 2019, doi: https://doi.org/10.1016/j.cities.2019.01.032
[8] R. Dias and A. Torkamani, "Artificial intelligence in clinical and genomic diagnostics," Genome Medicine, vol. 11, no. 1, Nov. 2019, doi: https://doi.org/10.1186/s13073-019-0689-8. Available: https://genomemedicine.biomedcentral.com/articles/10.1186/s13073-019-0689-8
[9] G. Schneider, "Automating drug discovery," Nature Reviews Drug Discovery, vol. 17, no. 2, pp. 97–113, Dec. 2017, doi: https://doi.org/10.1038/nrd.2017.232. Available: https://www.nature.com/articles/nrd.2017.232
[10] J. Vamathevan et al., "Applications of machine learning in drug discovery and development," Nature Reviews Drug Discovery, vol. 18, no. 6, pp. 463–477, Apr. 2019, doi: https://doi.org/10.1038/s41573-019-0024-5. Available: https://www.nature.com/articles/s41573-019-0024-5
[11] Sunkle, Sagar, et al. "Integrated "Generate, Make, and Test" for Formulated Products Using Knowledge Graphs." Data Intelligence, vol. 3, no. 3, 1 Jan. 2021, pp. 340–375, doi.org/10.1162/dint_a_00096, https://doi.org/10.1162/dint_a_00096. Accessed 19 Dec. 2024.
[12] A. Sparkes et al., "Towards Robot Scientists for autonomous scientific discovery," Automated Experimentation, vol. 2, no. 1, p. 1, 2010, doi: https://doi.org/10.1186/1759-4499-2-1
[13] B. Burger et al., "A mobile robotic chemist," Nature, vol. 583, no. 7815, pp. 237–241, Jul. 2020, doi: https://doi.org/10.1038/s41586-020-2442-2
[14] Z. Li, Y. Zhou, Q. Sheng, K. Chen, and J. Huang, "A High-Robust Automatic Reading Algorithm of Pointer Meters Based on Text Detection," Sensors, vol. 20, no. 20, p. 5946, Oct. 2020, doi: https://doi.org/10.3390/s20205946
[15] L. Wang, P. Wang, L. Wu, L. Xu, P. Huang, and Z. Kang, "Computer Vision Based Automatic Recognition of Pointer Instruments: Data Set Optimization and Reading," Entropy, vol. 23, no. 3, p. 272, Feb. 2021, doi: https://doi.org/10.3390/e23030272
[16] G. Bobovnik, T. Mušič, and J. Kutin, "Liquid Level Detection in Standard Capacity Measures with Machine Vision," Sensors, vol. 21, no. 8, p. 2676, Apr. 2021, doi: https://doi.org/10.3390/s21082676
[17] H. K. Regmi, J. Nesamony, S. M. Pappada, T. J. Papadimos, and V. Devabhaktuni, "A System for Real-Time Syringe Classification and Volume Measurement Using a Combination of Image Processing and Artificial Neural Networks," Journal of Pharmaceutical Innovation, vol. 14, no. 4, pp. 341–358, Oct. 2018, Doi: https://doi.org/10.1007/s12247-018-9358-5
[18] J. Redmon, S. Divvala, R. Girshick, and A. Farhadi, "You Only Look Once: Unified, Real-Time Object Detection," 2016 IEEE Conference on Computer Vision and Pattern Recognition (CVPR), pp. 779–788, 2016, doi: https://doi.org/10.1109/cvpr.2016.91. Available: https://arxiv.org/pdf/1506.02640.pdf
[19] OpenCV, "OpenCV library," Opencv.org, 2019. Available: https://opencv.org/. [Accessed: Jan. 14, 2025]
[20] R. Smith, "An Overview of the Tesseract OCR Engine," Ninth International Conference on Document Analysis and Recognition (ICDAR 2007) Vol 2, Sep. 2007, doi: https://doi.org/10.1109/icdar.2007.437699